\documentclass{article}
\usepackage{ijcai25}

\usepackage{times}
\usepackage{soul}
\usepackage{url}
\usepackage[hidelinks]{hyperref}
\usepackage[utf8]{inputenc}
\usepackage[small]{caption}
\usepackage{graphicx}
\usepackage{amsmath}
\usepackage{amsthm}
\usepackage{booktabs}
\usepackage{algorithm}
\usepackage{algorithmic}
\usepackage[switch]{lineno}
\usepackage[table]{xcolor}
\usepackage{array}
\usepackage{booktabs}
\usepackage{threeparttable}

\title{Federated Learning at the Forefront of Fairness: A Multifaceted Perspective}

\author{
Noorain Mukhtiar$^1$
\and
Adnan Mahmood$^1$\and
Yipeng Zhou$^{1}$\and
Jian Yang$^{1}$\and \\
Jing Teng$^{2}$\And
Quan Z. Sheng$^{1}$\\
\affiliations
$^1$School of Computing,  
Macquarie University, Sydney, Australia\\
$^2$School of Control and Computer Engineering,  
North China Electric Power University, China\\
\emails{
noorain.mukhtiar@hdr.mq.edu.au, \{adnan.mahmood, yipeng.zhou, jian.yang, michael.sheng\}@mq.edu.au}, {jing.teng@ncepu.edu.cn}}

\begin{document}

\maketitle

\begin{abstract}
Fairness in Federated Learning (FL) is emerging as a critical factor driven by heterogeneous clients' constraints and balanced model performance across various scenarios. 
In this survey, we delineate a comprehensive classification of the state-of-the-art fairness-aware approaches from a multifaceted perspective, i.e., \emph{model performance-oriented} and \emph{capability-oriented}. Moreover, we provide a framework to categorize and address various fairness concerns and associated technical aspects, examining their effectiveness in balancing equity
and performance within FL frameworks. We further examine several significant evaluation metrics leveraged to measure fairness quantitatively. Finally, we explore exciting 
open research directions and propose prospective solutions that could drive future advancements in this important area, laying a solid foundation for researchers working toward fairness in FL.     

\end{abstract}

\section{Introduction}
Federated Learning (FL), a quickly-emerging decentralized machine learning approach, enables collaborative model training across multiple participating devices (a.k.a. clients) while preserving data privacy \cite{ijcai2024p919}. Unlike traditional centralized learning methods, FL allows each participating device to train a model locally using its own data, thereby ensuring that the raw data never leave the device. Instead, only the locally trained model's updated parameters are sent to the server, wherein they are aggregated to constitute a global model. This, in turn, minimizes communication cost and transmission latency, and addresses several critical data issues, including but are not limited to, data access, security, and privacy \cite{ijcai2023_survey}.

Besides numerous advantages, the decentralized nature of FL introduces significant challenges, with {\em fairness} being the most critical factor. Conventional FL approaches primarily rely upon the random client selection scheme \cite{mcmahan2017communication,XiangLi} which can inadvertently exclude clients with less capabilities from the training process \cite{balakrishnan2022diverse}. Such approaches often leverage on threshold-based criteria, i.e., bandwidth availability \cite{Xu_bandwidth}, convergence speed \cite{convergencespeed}, utility \cite{OORT}, and local accuracy \cite{AAAImodelperformance}, in a bid to filter out less qualified clients and select the higher-quality ones, thereby introducing {\em bias} in FL systems. This biasing, if not addressed effectively, can pose a profound impact on both the server(s) and the client(s) sides in an FL environment. It is pertinent to mention that unfair treatment can provoke distrust and dissatisfaction in clients, thus discouraging their participation in the FL training process. Furthermore, treating all clients in an equal manner, regardless of their respective contributions, can reduce the servers' ability to attract high-quality clients. This, therefore, reduces the model's effectiveness and leads to less generalized models.

In addition to fair client selection, ensuring equitable model performance across diverse clients is equally crucial to the effectiveness of FL models. Clients with skewed or imbalanced data can disproportionately influence the predictions of a model, which can lead to suboptimal performance. It is worth mentioning that an unfair model can experience significant negative consequences, i.e., it may (a) fail to capture the true representation of the client population, (b) tend to under- or over-fitting, (c) marginalize certain clients, and (d) cause inefficient decision-making. Such disparities could reduce the effectiveness of FL models, discourage client participation, and hinder the overall progress of the FL system.

Ensuring fairness is not just a matter of equal treatment, but it involves addressing complex challenges that can hinder the long-term sustainability and effectiveness of FL systems. Integrating fairness into FL fosters an inclusive, robust, and efficient learning environment adaptable to diverse client capabilities. As research on fairness-aware FL is gaining significant momentum, 
several surveys have been published, which have primarily concentrated on client selection mechanisms and inherent biases introduced by data and device heterogeneity. For instance, \cite{Shi} provides an in-depth analysis of fairness-aware approaches in FL and discusses their related challenges. \cite{chen2023privacy} offers a comprehensive review of privacy and fairness challenges, their mitigation strategies, and trade-offs within FL environment. \cite{rafi2024fairness} presents a broad discourse on the interplay between privacy and fairness in FL. 
Despite these valuable contributions, prior works have predominantly framed fairness through the lens of client selection strategies, overlooking the equally critical dimension of model performance disparities. 

\begin{figure*} [!tb]
\centering
    \includegraphics[width=0.95\textwidth, trim = 11cm 10cm 11cm 10cm, 
    clip]{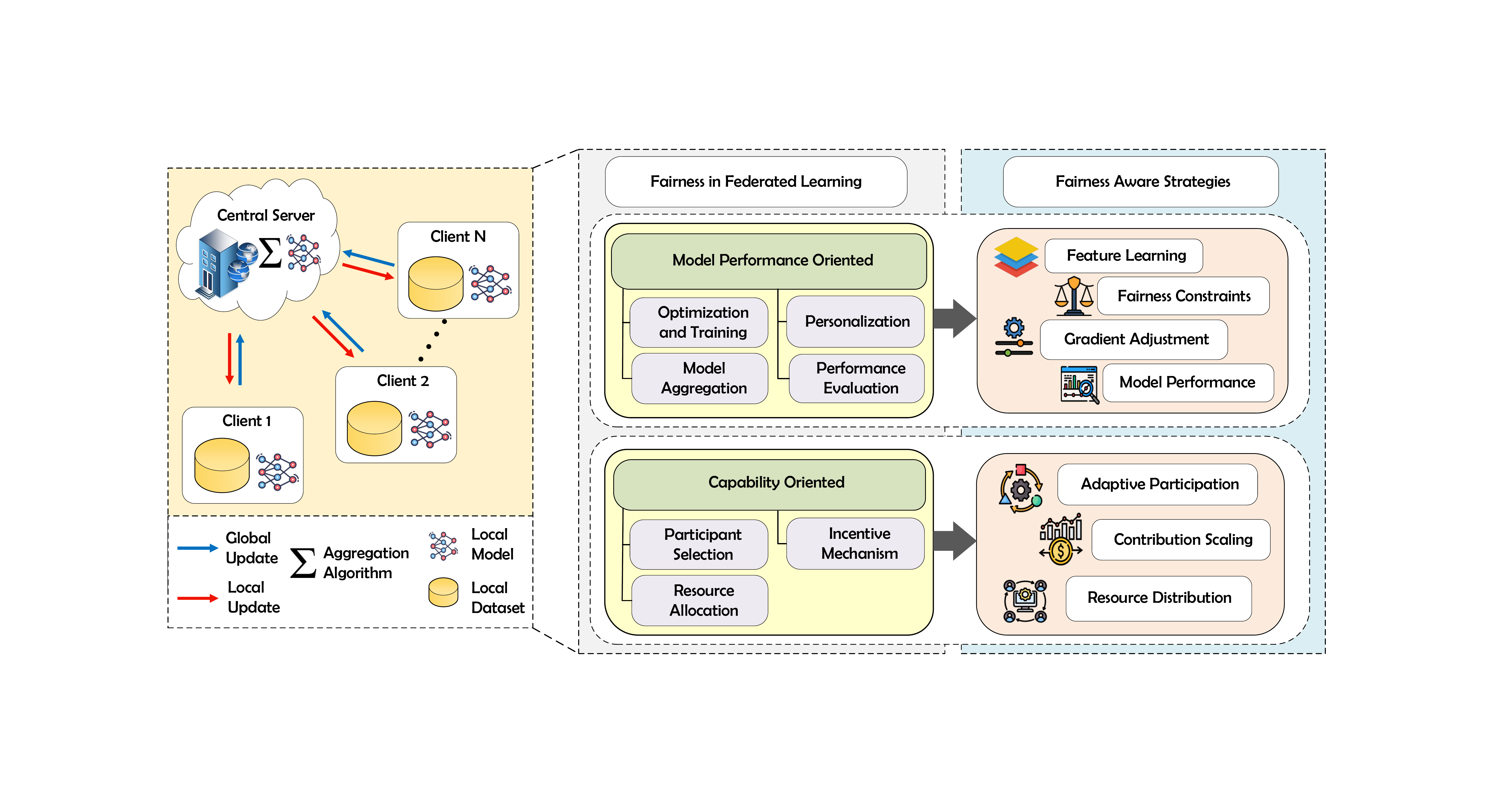}
    \caption{A taxonomy of fairness in federated learning, categorizing into model performance-oriented and capability-oriented approaches.}
    \label{Taxonomy}
\end{figure*}

Given the pressing need for a more holistic perspective, this survey systematically categorizes the state-of-the-art fairness-aware FL approaches, addressing both model performance and clients' capability considerations. To the best of our knowledge, this is the first survey to provide a structured and comprehensive analysis of fairness-aware strategies in FL, expanding the scope beyond client selection mechanisms. By providing a nuanced analysis of fairness considerations vis-\`a-vis model performance and clients' capability (illustrated in Figure \ref{Taxonomy}), this work aims to bridge existing gaps in the literature by offering a multifaceted perspective on the evolving landscape of fairness in FL systems.  Furthermore, we propose several promising future research directions that emphasize unresolved challenges and areas for exploration and set the stage for continued advancements in the field.
The main contributions of this survey are as follows:
\begin{itemize}
    \item We summarize various notions of fairness adopted in the state-of-the-art literature, highlighting their diverse foundations and conceptual frameworks within the field.
    
    \item We design a novel taxonomy of the fairness-aware FL from a multifaceted perspective, i.e., \emph{model-performance-oriented} and \emph{capability-oriented}, based on their key characteristics, underlying methodologies, and technical aspects.
    
    \item We outline key fairness evaluation metrics used by existing literature to offer a thorough insight into fairness quantification. 
    
    \item We identify significant challenges encountered while ensuring fairness and discuss open research directions by examining the state-of-the-art research in the field,
    highlighting existing limitations, and proposing future avenues for exploration.
\end{itemize}

The rest of the paper is organized as follows:
Section~\ref{SubSec: Fairness in FL} summarizes the fundamental interpretations of fairness and delves into model performance-oriented and capability-oriented approaches in FL.
Section \ref{Sec: Fairness Evaluation} evaluates the adoption trends of fairness evaluation metrics in existing FL research. 
 Section \ref{Sec: Open Research Directions} identifies
 open research directions within the realm of bias mitigation in FL systems. Finally, Section \ref{Conclusion} provides some concluding remarks.

\section{Fairness in Federated Learning} 
\label{SubSec: Fairness in FL}
FL encounters unique challenges and opportunities regarding fairness. In this section, we will discuss the broad notions of fairness in terms of FL frameworks, their underlying principles, and associated techniques.

\subsection{Fairness Notions in Federated Learning}
\label{SubSec: Fairness notions in FL}
\textit{Client-level Fairness} \cite{Shi} ensures equitable participation of all clients involved in the FL process. This notion aims to mitigate potential disparities by prioritizing the inclusion of underrepresented or unrepresented participants. 
\textit{Group Fairness} \cite{Wei,ezzeldin} maintains equity among different demographic groups of clients participating in federated training. This concept focuses on alleviation of biases in the trained model's performance against specific groups based on sensitive attributes, i.e., age, gender, and race. 
\textit{Performance Distribution Fairness} \cite{rafi2024fairness} measures the uniformity of performance distribution across FL clients. The main idea is to achieve a correct balance between fairness and accuracy while ensuring comparable performance levels for each client.
 
\textit{Good-Intent Fairness} \cite{li2020fair_ICLR,mohri2019agnostic} aims to reduce the maximum loss among protected groups, thereby preventing the overfitting of any single model at the cost of others. \textit{Contribution Fairness} \cite{Lingjuan,yu} ensures the distribution of rewards in accordance with client's contribution, i.e., uploaded parameters or gradients. This notion prioritizes rewarding clients with the maximum contribution in FL training rather than those contributing negligibly. \textit{Regret Distribution Fairness} \cite{rafi2024fairness,yu} reduces the discrepancy among FL clients' regrets on waiting times to obtain incentive payouts, considering the duration a client has waited to receive full payoff. \textit{Expectation Fairness} \cite{Shi} aims to establish fairness by minimizing the inequality among clients over time as incentive rewards are incrementally disbursed. Each of these mentioned fairness notions brings forth unique advantages across diverse FL scenarios. 

Fairness-oriented approaches require striking a balance between the interests of both FL clients and the server by emphasizing the inclusion of disadvantaged clients and mitigation of model performance bias to produce more representative and unbiased models. These approaches can be broadly categorized into {\em model performance-oriented} and {\em capability-oriented}. Understanding the unique characteristics and implications of each technique is essential for selecting an appropriate fairness-aware scheme in FL.  

\subsection{Model Performance-Oriented Approaches}
\label{Sub_sec: Model Performance Oriented Approaches}

Model performance-oriented fairness approaches in FL prioritize ensuring equitable predictive performance across all clients, regardless of variations in data distribution, class imbalance, participation frequency, and computational resources. These approaches address fairness (via either one or a combination of) formulating \emph{optimizing objective}, designing \emph{personalized models}, employing \emph{aggregation techniques}, and incorporating \emph{fairness-aware evaluation} metrics. By integrating fairness constraints, these techniques strive to balance accuracy, robustness, and fairness among participating entities to mitigate systematic biases that could worsen model performance for a certain subset of clients. The subsequent discussion elaborates on key performance-oriented approaches in a detailed manner.

\subsubsection{Fairness as an Optimization Problem}
\label{Subsub_sec: Fairness as Optimization Problem}
An optimization based strategy is adopted to tackle bias-related challenges encountered during FL model training. It involves the formulation of (either single- or multi-objective) local/global optimization problem while satisfying the target fairness constraints.

FedISM \cite{IJCAI_24_Fairness} incorporates sharpness-aware local optimization, wherein each client simultaneously minimizes both loss and sharpness during training, striving to harmonize sharpness levels across clients for fair generalization. Moreover, it employs a sharpness-dependent global aggregation strategy to weight client updates based on their respective sharpness levels. This method prioritizes uniform sharpness distribution across clients, thereby enhancing the fairness in model performance. Similarly, FedLF \cite{FedLF} formulates a multi-objective optimization problem incorporating an effective fairness-driven objective for FL. The algorithm aims at mitigating layer-level gradient conflicts by computing layer-wise direction fragments while reducing improvement bias.  Subsequently, optimal layer-wise update directions are determined and concatenated to form a unified model update, thereby preventing domination by a single client's gradient.

\subsubsection{Fairness-aware Model Aggregation}
\label{Subsub_sec: Fairness in Model Aggregation}
The fairness-aware aggregation strategy is employed during the model aggregation step to ensure equitable client contributions and mitigate the risk of dominance by specific participants. It balances the influence of each client on the aggregated model by using various methods, i.e.,  weighted aggregation, adaptive client reweighting, and gradient alignment.

FedHEAL \cite{chen2024fairCvpr} proposes to address performance fairness in FL under domain skew by resolving parameter update conflicts and model aggregation bias. Initially, the algorithm discards insignificant parameter updates based on discovered characteristics to avoid poorly performing clients from being overwhelmed by those updates. Subsequently, a fair aggregation objective is incorporated to prevent the global model bias towards certain domains, ensuring the continuous alignment of the global model with an unbiased model. 
Another fairness-aware aggregation algorithm FairFed \cite{ezzeldin} is designed to adjust clients’ weights during global aggregation while minimizing Statistical Parity Difference (SPD) and Equal Opportunity Difference (EOD). This is based on local debiasing, where clients assess the fairness of the global model on their local datasets in each round and collaborate with the server to adjust aggregation weights based on the mismatch between the global and local fairness measurements, i.e., gap metric, favoring clients whose local metrics align with the global measure. 

\subsubsection{Fairness via Model Personalization}
\label{Subsub_sec: Fairness via Model Personalization}
Fairness via model personalization \cite{Collab_KDD_22,CVPR_Personlization} is leveraged to tailor models to clients' specific data characteristics. As diverse clients may have distinct learning objectives, a one-size-fits-all model can cause performance disparities. Therefore, personalization  techniques \cite{LiuZ0HGS24} aim to adapt models based on clients with similar data distributions for fostering more effective knowledge transfer.

The research in \cite{IJCAI_Shapley} envisages the ShapFed-WA approach to foster contribution fairness by personalizing participants' updates based on their contributions. Such contributions are assessed leveraging Shapley values (SV) to provide a detailed understanding of class-specific influences. DBE \cite{NEURIPS2023_2e0d3c6a} addresses representation bias and degeneration phenomena by bidirectional knowledge transfer between the server and clients to reduce domain discrepancies in the representation space. The framework is composed of two key modules. The first module detaches the representation bias from the original representations and stores it in a personalized representation bias memory on each client. In the subsequent module, mean regularization guides local feature extractors to extract representations with a consensual global mean during local training. This, in turn, eliminates conflicts between personalized and generic features, improving representation quality in the shared space.

\subsubsection{Fairness in Performance Evaluation}
\label{Sub_sec: Fairness in Performance Evaluation}
Fair performance evaluation approaches are implemented to provide a more equitable and nuanced evaluation of FL models' performance, addressing the complexities introduced by varying client conditions. By incorporating fairness-aware performance metrics, the evaluation process ensures that all clients, regardless of their disparities, contribute equitably to the global model's performance.

EAB-FL~\cite{ijcai2024_EABFL} introduces the model poisoning attack, targeting group unfairness in FL systems by allowing malicious clients to upload poisoned model updates that intentionally exacerbate bias against specific demographic groups while maintaining overall model utility. To intensify model bias, each malicious client identifies a subset of their local training samples negatively impacting the performance of specific demographic group (the targeted group). Furthermore, an influence score is utilized to measure the effect of each local training sample on the model’s performance, which evaluates how the model predictions related to the targeted group would be altered if a particular training samples are excluded from the dataset, particularly under the fairness constraints imposed on the classifier.
AGLFOP \cite{hamman2024demystifying_ICLR24} 
is an algorithm based on partial information decomposition, identifying three main sources of unfairness in FL, i.e., Unique Disparity, Redundant Disparity, and Masked Disparity. The fundamental limits on the trade-off between global and local fairness 
are established using decomposition, highlighting their alignment or conflict. Moreover, a convex optimization problem, i.e., accuracy and global-local fairness optimality, is formulated for defining the theoretical boundaries of the accuracy-fairness trade-off.

\subsection{Capability-Oriented Approaches}
\label{Sub_sec: Capability-Oriented}
Capability-oriented fairness approaches emphasize reducing disparities in clients' ability to contribute effectively to the FL process. In practical scenarios, clients vary in computational power, data availability, and communication capabilities, leading to biased training. These approaches address the inherent disparities by leveraging some strategies, i.e., \emph{adaptive client selection}, \emph{fair resource distribution}, and \emph{contribution scaling}, assuring that clients with lower capabilities are not unfairly excluded from FL training. By fostering equitable participation, these approaches enhance fairness and prevent dominant clients from disproportionately influencing the global model. The subsequent discussion delves into various approaches designed to uphold fairness from a participants' capability perspective.

 \subsubsection{Fair Participant Selection}
\label{Subsub_sec: Fair Participant Selection}
A fair client selection strategy is used to promote fairness in the process of choosing clients for FL training, aiming to mitigate biases that could negatively impact the model’s performance for individual or specific group of clients. Fair client selection involves two stages, namely (a) choosing \textit{initial client pool} for each FL task and (b) selecting \textit{subset of clients} for each FL round. It is pertinent to mention that even if a client is chosen to participate in FL training, it does not guarantee its participation in every FL round due to several factors, i.e., communication and computation cost, conflicting schedule, low battery, or unstable network connectivity.

FairFedCS \cite{shi2023fairness} 
focuses on maintaining a balance between model performance and fairness considerations in FL client selection. The proposed algorithm is based on the Lyapunov optimization problem, designed to dynamically adjust FL client selection probabilities aligned with their respective participation frequency, reputation, and contribution to the model performance. This approach neglects reputation threshold filtering to promote fair treatment and clients are allowed to restore their reputations despite achieving poor performance.
RBCS-F \cite{huang2020efficiency} 
introduces a reputation-based participant selection algorithm based on the Contextual Combinatorial Multi-Armed Bandit \begin{math} (C^2MAB)\end{math} method to estimate each client's model exchange time as per their historical reputation.  
This work concentrates on minimizing the average model exchange time between the server and each client while adhering to relatively flexible long-term fairness and system constraints. The fairness challenge is formulated as a Lyapnov optimization problem to improve FL clients' participation rates. Moreover, penalty factor is specified to balance the trade-off between the objective function minimization and fairness constraint satisfaction. 

\begin{table*}[!tb] 
    \centering
    \scriptsize
    \definecolor{softblue}{RGB}{173, 216, 230}
    \begin{threeparttable}
    \begin{tabular}{>{\raggedright\arraybackslash}p{1.7cm}
>{\raggedright\arraybackslash}p{2.2cm}
>{\raggedright\arraybackslash}p{2.8cm}
>{\raggedright\arraybackslash}p{3.2cm}
>{\raggedright\arraybackslash}p{1cm}
>{\raggedright\arraybackslash}p{4cm}}
        \toprule
\multicolumn{1}{c}{\textbf{Method}} &
\multicolumn{1}{c}{\textbf{Fairness Notion(s)}} &
\multicolumn{1}{c}{\textbf{Associated Algorithm(s)}} &
\multicolumn{1}{c}{\textbf{Datasets}} &
\multicolumn{1}{c}{\textbf{Venue}} &
\multicolumn{1}{c}{\textbf{Key Idea}} \\[-1.5pt]
\midrule
\addlinespace[1pt]
       \rowcolor{softblue} 
        \multicolumn{6}{c}{\textbf{Model Performance-Oriented Approaches}} \\ [-2pt]
        \midrule
        ShapFed-WA\textsuperscript{[1]} & Collaborative Fairness & Weighted Aggregation and Personalization & CIFAR-10, Chest X-ray, and Fed-ISIC 2019 & IJCAI`24 & Fine-grained evaluation of participant contributions \\
\addlinespace[2pt]
FedLF\textsuperscript{[2]} & Individual Fairness & Multi-objective Optimization with Fairness-driven Objective  & Fashion-MNIST, CIFAR-10, and CIFAR-100  & AAAI`24 & Layer-wise fair direction calculation to mitigate the improvement bias \\      \addlinespace[2pt] FedISM\textsuperscript{[3]} & Individual and Performance Distribution Fairness & Inter-client Sharpness Matching & RSNA ICH and ISIC 2019 & IJCAI`24 & Weighting client updates based on their respective sharpness levels \\      FedHEAL\textsuperscript{[4]} & Performance Distribution Fairness & Parameter Update Consistency & Digits and Office-Caltech & CVPR'24 & Discarding unimportant parameter updates to prevent disparities\\ \addlinespace[2pt]
        EAB-FL\textsuperscript{[5]} &  Group Fairness & Optimization with Fairness Constraint & CelebA, Adult Income, UTKFaces, and MovieLens 1M & IJCAI`24 & Exacerbating group unfairness by launching model poisoning attack \\ \addlinespace[2pt]
    AGLFOP\textsuperscript{[6]} & Global and Local Fairness & Partial Information Decomposition & Synthetic and Adult & ICLR`24 & Optimizing the joint distributions of model predictions while constraining both global and local fairness metrics \\       \addlinespace[2pt]
    FairFed\textsuperscript{[7]} & Group Fairness & Fairness-aware Aggregation  & Adult, COMPAS, ACSIncome, and TILES  & AAAI`23 & Fairness-aware aggregation to enhance group fairness \\ \addlinespace[1pt]
         DBE\textsuperscript{[8]} & Performance Distribution Fairness & Domain Bias Eliminator  & Tiny-ImageNet, CIFAR-100, Fashion-MNIST, and AG News  & NeurIPS`23 & Promoting the bi-directional knowledge transfer between server and clients \\
        \midrule
        \addlinespace[1pt]
         \rowcolor{softblue} 
        \multicolumn{6}{c}{\textbf{Capability-Oriented Approaches}} \\ [-2pt]
        \midrule 
         Rank-Core-Fed\textsuperscript{[9]} & Good Intent Fairness & Proportional Veto Core & MNIST and CIFAR-10 & ICML'24 & Measuring output models' quality based on ordinal rank instead of cardinal utility \\ \addlinespace[3pt]        FairFedCS\textsuperscript{[10]} & Contribution Fairness & Lyapnov Optimization & MNIST and CIFAR-10 & ICME'23 & Adjusting client selection
         probabilities based on reputation, participation frequency, and contribution to the
         model \\        \addlinespace[1pt]
         FEEL\textsuperscript{[11]} & Performance Distribution Fairness & Client Clustering and Greedy Selection & FEMNIST and CIFAR-10 & IEEE TNSM'23 & Two-phased client selection and scheduling \\ 
         \addlinespace[3pt]
         RRAFL\textsuperscript{[12]} & Contribution Fairness & Reputation based Incentive Mechanism  & MNIST, Fashion MNIST, and IMDB & WWW'21 & Incentive mechanism based on reputation and reverse auction theory \\        \addlinespace[2pt]  qFFL\textsuperscript{[13]} & Performance Distribution Fairness &  Fair Resource Allocation & Synthetic, Vehicle, Sent140, and
        Shakespeare & ICLR'20 &
Assigning weights to devices as per their losses \\ \addlinespace[3pt]
        RBCS-F\textsuperscript{[14]} & Performance Distribution Fairness & Lyapunov Optimization and $C^2MAB$ & Fashion-MNIST and CIFAR-10 & IEEE TPDS'20 & Estimating each client’s model exchange time based on its
historical reputation \\
        \bottomrule
    \end{tabular}
     \begin{tablenotes} \centering
            \item Note: \textsuperscript{[1]}\cite{IJCAI_Shapley}; \textsuperscript{[2]}\cite{FedLF}; \textsuperscript{[3]}\cite{IJCAI_24_Fairness}; \textsuperscript{[4]}\cite{chen2024fairCvpr}; \textsuperscript{[5]}\cite{ijcai2024_EABFL};
            \textsuperscript{[6]}\cite{hamman2024demystifying_ICLR24};\\
            \textsuperscript{[7]}\cite{ezzeldin}; \textsuperscript{[8]}\cite{NEURIPS2023_2e0d3c6a}; \textsuperscript{[9]}\cite{rankcorefed2024fair}; \textsuperscript{[10]}\cite{shi2023fairness}; \textsuperscript{[11]}\cite{albaseer2023fair}; \textsuperscript{[12]}\cite{Incentive_WWW21}; \\
            \textsuperscript{[13]}\cite{li2020fair_ICLR}; \textsuperscript{[14]}\cite{huang2020efficiency}.
        \end{tablenotes}
    \end{threeparttable}
    \caption{Comparison of the state-of-the-art fairness-oriented approaches in federated learning.}
    \label{table_1}
\end{table*}
\subsubsection{Fair Resource Allocation}
\label{subsub_sec: Fair Resource Allocation}
A fair resource allocation strategy is adopted to ensure equitable resource distribution across FL participants encompassing heterogeneous specifications, i.e., computational capacity, data quality, and network limitations.

FEEL \cite{albaseer2023fair} presents a dual-phased participant selection and scheduling scheme for clustered multi-task FL, focusing on the improvement of model convergence speed while considering all data distribution. In the initial phase, an algorithm maintains correct clustering and fairness across participants by leveraging bandwidth reuse for slower clients (those who take longer time to train their models) and utilizing device heterogeneity to schedule participants based on their delays. In the second phase, the server clusters participants as per the pre-determined threshold values and stopping conditions. When a client cluster meets the stopping criteria, the server uses a greedy algorithm to select clients with better resources and lower delay. Another work $q$-FFL \cite{li2020fair_ICLR}, motivated by fair resource distribution in wireless networks, proposes an optimization objective to encourage fair and uniform accuracy distribution among FL devices. The algorithm minimizes a combined reweighted loss characterized by parameter $q$, which allocates greater weights to the devices with greater losses and vice versa.

\subsubsection{Fair Incentive Mechanism}
\label{subsub_sec: Fair Incentive Mechansim}

A fairness-driven incentive mechanism is implemented to align client motivations with the overall goals of FL system, fostering equitable participation and encouraging clients to contribute meaningfully to the model. Common techniques include reputation-based, reward distribution-based, regret-based, and resource-aware incentives \cite{YangZHWS23}. 

Rank-Core-Fed \cite{rankcorefed2024fair} investigates a notion of Proportional Veto Core (PVC) to ensure fairness of the utility distribution among the participating agents. The algorithm aims to achieve fairness by ensuring that the final model is PVC-stable, considering the ordinal preferences of agents. It guarantees fairness based on the models' ordinal rank rather than solely relying on their utility values.  
RRAFL \cite{Incentive_WWW21} introduces a marginal contribution-based model quality detection algorithm coupled with client contribution evaluation, which utilizes the reputation and reverse auction theory. Initially, participants bid for tasks with reputation serving as an indirect measure of their reliability and
data quality. Subsequently, they are selected and rewarded
based on both their reputation and bids, while adhering to a
limited budget.

While our taxonomy distinctly distinguishes model performance-oriented and capability-oriented fairness approaches, it is pertinent to mention that these categories are not mutually exclusive and may overlap in certain scenarios. For instance, some capability-oriented methods, i.e., fair incentive mechanisms,
can inherently impact model performance by encouraging broader participation, reducing client dropout, and improving data heterogeneity.

\section{Fairness Evaluation}
\label{Sec: Fairness Evaluation}
Fairness, being a concept largely premised on ethics and the social choice theory, remains challenging to evaluate, particularly in the context of the FL systems. Delineating a robust set of performance evaluation measurements is essential to ensure the validity of fairness-aware FL algorithms. In this section, we explore various fairness evaluation metrics and their adoption trends tailored to specific FL scenarios.

\subsection{Average Variance and Standard Deviation} 
\label{SubSec: Average Variance}
Average Variance (AV) evaluates the variability of a set of values over time $t$ to quantify how much individual data points differ from the overall average. 
In the context of fairness in FL, it quantifies the level of fairness during model optimization, where a given model is considered fairer if its variance is less than others. \cite{wang2021FedFV} adopts average and variance to assess the performance disparities. Mathematically, AV is written as Equation (\ref{Average Variance Eq}):      
\begin{equation} \label{Average Variance Eq}
    AV = \frac{1}{n} \sum_{i=1}^{n} (F_{i}(t) - \bar{F}(t))^2 
\end{equation}
The work in \cite{chen2024fairCvpr,zhao2022participant} uses the standard deviation (SD) of performance across participants to quantify fairness. SD is expressed as Equation (\ref{Standard Dev Eq}):
\begin{equation} \label{Standard Dev Eq}
{
\sigma = \sqrt{\frac{1}{n} \sum_{i=1}^{n} (F_{i}(t) - \bar{F}(t))^2}
}
\end{equation}
where $F(t)$ represents the accuracy of a model on client $i$ at time $t$, and $\bar{F}(t)$ is the average accuracy across all $n$ clients. 

Since both SD and AV quantify the variability within a distribution, they share the fundamental concept of measuring dispersion. Mathematically, they are closely related as variance provides the squared measure of deviation from the mean, whereas, SD refines this by taking the square root to maintain the original unit of measurement.

\subsection{Pearson Correlation Coefficient} 
\label{SubSec: Pearson Correlation Coefficient}
The Pearson Correlation Coefficient (PCC) measures the strength and direction of the linear relationship between two variables. In the context of SV and contribution assessment methods, PCC calculates how closely the predicted contribution values align with the true SVs. 
In \cite{shi2022fedfaim}, the authors apply PCC with true SV as an assessment metric for contribution evaluation. A higher PCC value indicates greater fairness in assessing the client's contribution to the FL model. PCC is expressed as Equation (\ref{Pearson Correlation Coefficient Eq}):
\begin{equation} \label{Pearson Correlation Coefficient Eq}
       PCC = \frac{\sum_i (\phi^*_i-\overline{\phi^*}) (\phi_i-\overline\phi)}{S_{\phi^*_i} \times {S_{\phi_i}}}    
\end{equation} 
where $\phi_i$ represents the SV for client $i$, $\phi^*_i$ shows the ground-truth SV, 
$\overline{\phi^*}$ and $\overline\phi$ are mean values of $\phi^*$ and $\phi$ respectively. $S_{\phi^*_i}$ and ${S_{\phi_i}}$ denote relative SDs.

\subsection{Jain’s Fairness Index}
\label{SubSec: Jain’s Fairness Index}
Jain’s Fairness Index (JFI) metric evaluates fairness in resource allocation among $n$ participants. The work in \cite{shi2023fairness} adopts JFI to assess the level of fairness achieved after model convergence. JFI is computed as Equation (\ref{Jain’s Fairness Index Eq}):
\begin{equation} \label{Jain’s Fairness Index Eq}
    JFI = \frac{(\sum_{i=1}^n F_i (t))^2}{n.\sum_{i=1}^n (F_i (t))^2} 
\end{equation} 
where $F_i (t)$ represents local objective function of client $i$. The values of JFI  range from $0$ indicating highly unfair to $1$ representing most fair.
    
\subsection{Statistical Parity Difference}
\label{subSec: Statistical Parity Difference}
Statistical Parity Difference, also referred to as Demographic parity, quantifies the difference in positive outcomes across various demographic groups. It is calculated as Equation (\ref{Statistical Parity Difference Eq}):
\begin{equation}  \label{Statistical Parity Difference Eq}
    SPD = |P(\hat{Y} = 1|A = 0) - P(\hat{Y} = 1|A = 1)|
\end{equation}
herein $P(\hat{Y} = 1|A = 0)$ indicates probability of a positive outcome for unprivileged group, and $P(\hat{Y} = 1|A = 1)$ denotes probability of a positive outcome for privileged group. 

\subsection{Equal Opportunity Difference} 
\label{SubSec: Equal Opportunity Difference}
Equal Opportunity Difference assesses the performance of a binary predictor $\hat{Y}$ concerning the sensitive attribute $A$ and the actual outcome $Y$. A predictor is considered to be fair if the true positive rate is not influenced by the sensitive attribute $A$. It can be computed by Equation (\ref{EoD Eq}):
\begin{equation} \label{EoD Eq}
    \resizebox{0.91\linewidth}{!}{$
            \displaystyle
              EOD = P(\hat{Y} = 1|A = 0,Y=1) - P(\hat{Y} = 1|A = 1, Y=1)
             $}
   \end{equation}
where $P(\hat{Y} = 1|A = 0,Y=1)$ denotes the probability of a positive outcome for unprivileged group, and $P(\hat{Y} = 1|A = 1, Y=1)$ indicates the same for a privileged group.  

The work in \cite{ezzeldin,hamman2024demystifying_ICLR24} utilizes SPD and EOD as fairness evaluation metrics to measure the performance of the envisaged fair client selection algorithm and demonstrate its effectiveness in addressing bias in FL scenarios. The values closer to zero suggest greater fairness and positive values in these metrics indicate that the unprivileged group outperforms the privileged group.

\section{Open Research Directions}
\label{Sec: Open Research Directions}
Despite extensive research conducted in recent years, as thoroughly discussed in Section \ref{SubSec: Fairness in FL}, fairness in an FL environment still encounters
many open challenges. As FL continues to evolve, ensuring fairness across diverse clients considering key aspects, i.e., accuracy, privacy, model generalization, and utility, remains an intricate challenge and
requires further exploration. 
In this section, we elaborate on the crucial open research directions in FL to explore how each challenge interrelates with fairness and the potential trade-offs involved. By discussing the above mentioned aspects in detail, we provide insights into the state-of-the-art research landscape and identify prospective future research areas to create robust, fair, and high-performance FL models.

\subsection{Balancing Fairness and Accuracy}
\label{SubSec: Balancing Fairness and Accuracy}
A fair but less-performing model is not ideal, emphasizing the significance of balancing the trade-offs between accuracy and fairness in the FL environment. Ensuring fairness tends to maintain equitable outcomes across participants especially those with less representative data or limited resources. However, improving fairness may compromise the model accuracy, presenting an open research challenge. The research in \cite{lewis2024ensuring} highlights the interrelation between fairness and model accuracy, demonstrating a decline in overall model accuracy while improving fairness and vice versa.

The work in \cite{Changjian} strives to maintain a balance between accuracy and fairness on multiple clients' subgroups by introducing a bi-level optimization algorithm-based fair predictor. The lower-level subgroup-specific predictors are trained on limited data, while the upper-level fair predictor is adjusted to align with all subgroup-specific predictors. Whilst the empirical evaluations indicate the improvement in fair predictor without sacrificing accuracy, the method is bounded by certain conditions, i.e., the assumption of similar ground truth predictors (A-Bayes predictors) across different subgroups that could lead to non-trivial scenarios upon violation. This, in turn, presents a gap for the development of adaptive strategies required to maintain fairness and achieve competitive model performance. A possible research direction is to design fairness-aware loss function that incorporate differentiable fairness constraints directly into global model optimization to enable joint optimization without decoupling fairness from accuracy.
 
\subsection{Fairness and Privacy Trade-offs}
\label{SubSec: Fairness and Privacy Tradeoffs}
Ensuring fairness in FL can heighten privacy risks, as it frequently requires collecting clients' sensitive demographic information that may be extraneous to the related task. This gathered information is utilized to navigate model adjustments and reduce bias, nevertheless, increasing the potential for privacy breaches or exposure to sensitive data. The work in \cite{chang2021privacy} indicates that fair models increase privacy risks for underprivileged subgroups.

Fairness-aware models necessitate consistent performance across all subgroups, however, limited data for underprivileged groups can result in overfitting to the training data of privileged groups, thereby raising privacy concerns.  
\cite{esipova2023disparate} proposes a solution to this issue by considering cross-model fairness (where the cost of integrating privacy to a non-private model should be equitably distributed among different groups).
They explore gradient misalignment as a key factor in disparate impact within differentially private stochastic gradient descent and envisage global scaling method to mitigate it. The empirical results demonstrate that the proposed
method improves fairness in terms of accuracy and loss parameters without requiring protected groups' data and reduces disparate impact for all groups. However, lacks in fully eliminating biases from data collection or modeling assumptions, making independent fairness validation necessary for models with global scaling to prevent unintended disparities. Ultimately, maintaining fairness across all clients while respecting individual clients' privacy remains in its infancy, entailing innovative approaches that balance these competing priorities for trustworthy and privacy-preserving FL systems. A potential direction is to develop fairness-preserving noise calibration techniques that adjust privacy noise based on local training dynamics, e.g., gradient confidence or convergence behavior, instead of applying uniform noise across all clients or updates.
\subsection{Navigating Trade-off Between Fairness and Generalization}
\label{SubSec: Fairness and Generalization Tradeoffs}
A model achieving higher degree of fairness at the cost of generalization is not ideal for long-term sustainability in the FL environment. 
Model generalization refers to the ability of a model to perform well on unseen data. Maintaining fairness requires equitable model performance across diverse client data sources, which creates a conflict in maintaining broad generalization. 
For instance, achieving generalization often requires the model to learn broad patterns from diverse data. However, overemphasizing fairness may lead to overfitting, which, in turn, reduces the ability of the models to generalize well across other unseen data, hence compromising the models' broader applicability. 

The research in \cite{mohri2019agnostic} aims to achieve fairness by preventing the model from overfitting to any specific client at the expense of others. The global model is optimized for a mixed-client target distribution, while ensuring that the worst-performing client's loss does not increase. However, this approach only performs well with a small number of clients, and generalization becomes challenging as the client pool grows. The research in \cite{NEURIPS2019} envisages a solution to the above-mentioned challenge by designing an oracle-efficient algorithm for the fair empirical risk minimization task. The empirical evaluations demonstrate the effectiveness of the algorithm. Nevertheless, ensuring fairness across both new individuals and classification tasks requires a large number of samples, which can be difficult to obtain. Therefore, the challenge lies in designing algorithms that ensure fairness while preserving model generalization across heterogeneous clients in a bid to develop fair and resilient FL environments. A prospective solution is to explore adaptive regularization schemes that penalize fairness-induced overfitting during training by monitoring generalization bounds across cross-validation splits of client data.

\subsection{Bridging Gap Between Fairness and Utility}
\label{SubSec: Fairness and Utility Tradeoffs}
Balancing fairness and utility presents a critical yet challenging open research direction in FL. Utility focuses on maximizing the overall system performance, e.g., 
accuracy, efficiency, or convergence, while fairness aims to prevent biases by involving adjustment techniques to the training process that could affect model utility. Several studies have demonstrated that improving fairness can reduce utility, and vice versa. For instance, the research in \cite{Dehdashtian_2024_CVPR} delves into the inherent trade-offs between utility and fairness, providing several insights into how improving fairness may impact utility.

To address this, \cite{ACM2024privacy} offers a solution to achieve an optimal balance between fairness, utility, and privacy in FL systems.
This approach employs a fairness-aware optimization strategy by constraining model updates within a predefined fairness-preserving region. It utilizes confined gradient descent (CGD) to enforce a bounded fairness constraint, limiting the deviation between individual client models and the aggregated global model to prevent the system from disproportionately favoring dominant clients during model aggregation. The empirical results demonstrate that CGD significantly reduces accuracy variance across participants and outperforms baseline methods, e.g., 
FedAvg \cite{mcmahan2017communication} and Ditto \cite{li2021dittofairrobustfederated} in terms of fairness. 
However, it strictly relies on certain theoretical bounds and convergence guarantees, which could restrict its adaptability to highly non-convex loss functions and struggle in extreme data heterogeneity scenarios, thus leading to suboptimal utility. Accordingly, the complexity of ensuring fairness while maintaining utility in highly heterogeneous environments continues to be an open research challenge. A promising direction is to design dynamic fairness-aware utility maximization framework that adaptively modulates each client's influence during aggregation based on both of its marginal utility contribution and fairness deviation over time.

\section{Conclusion}
\label{Conclusion}
With the widespread adoption of Federated Learning (FL) in cutting-edge technologies, ensuring fairness has become a critical concern across diverse client populations. This survey delineates the multifaceted dimensions of fairness in FL, ranging from theoretical notions to practical implementations. We summarize and categorically segregate the state-of-the-art fairness-aware strategies based on the techniques utilized and examine the issues addressed in a bid to offer a detailed understanding of implications pertinent to fairness-aware FL. We discuss the evaluation metrics extensively utilized in literature to assess the performance of fairness-aware algorithms, aiming to enhance the robustness and sustainability of the FL environment.
We also identify several key areas for future research on this important topic,  paving a road map for researchers to shape a fairer future for FL.


\begin{thebibliography}{}

\bibitem[\protect\citeauthoryear{Albaseer \bgroup \em et al.\egroup }{2023}]{albaseer2023fair}
Abdullatif~Mohammed Albaseer, Mohamed Abdallah, Ala Al-Fuqaha, Abegaz~Mohammed Seid, Aiman Erbad, and Octavia~A Dobre.
\newblock {Fair Selection of Edge Nodes to Participate in Clustered Federated Multitask Learning}.
\newblock {\em IEEE Transactions on Network and Service Management}, 20(2):1502--1516, 2023.

\bibitem[\protect\citeauthoryear{Balakrishnan \bgroup \em et al.\egroup }{2022}]{balakrishnan2022diverse}
Ravikumar Balakrishnan, Tian Li, Tianyi Zhou, Nageen Himayat, Virginia Smith, and Jeff Bilmes.
\newblock {Diverse Client Selection for Federated Learning via Submodular Maximization}.
\newblock In {\em {International Conference on Learning Representations}}, pages 1--18, 2022.

\bibitem[\protect\citeauthoryear{Chai \bgroup \em et al.\egroup }{2021}]{convergencespeed}
Zheng Chai, Yujing Chen, Ali Anwar, Liang Zhao, Yue Cheng, and Huzefa Rangwala.
\newblock {FedAT: A High-Performance and Communication-efficient Federated Learning System with Asynchronous Tiers}.
\newblock In {\em {International Conference for High Performance Computing, Networking, Storage and Analysis}}, pages 1--16, 2021.

\bibitem[\protect\citeauthoryear{Chang and Shokri}{2021}]{chang2021privacy}
Hongyan Chang and Reza Shokri.
\newblock {On the Privacy Risks of Algorithmic Fairness}.
\newblock In {\em {IEEE European Symposium on Security \& Privacy}}, pages 292--303, 2021.

\bibitem[\protect\citeauthoryear{Chaudhury \bgroup \em et al.\egroup }{2024}]{rankcorefed2024fair}
Bhaskar~Ray Chaudhury, Aniket Murhekar, Zhuowen Yuan, Bo~Li, Ruta Mehta, and Ariel~D Procaccia.
\newblock {Fair Federated Learning via the Proportional Veto Core}.
\newblock In {\em {International Conference on Machine Learning}}, pages 42245--42257, 2024.

\bibitem[\protect\citeauthoryear{Chen \bgroup \em et al.\egroup }{2023}]{chen2023privacy}
Huiqiang Chen, Tianqing Zhu, Tao Zhang, Wanlei Zhou, and Philip~S Yu.
\newblock {Privacy and Fairness in Federated Learning: On the Perspective of Tradeoff}.
\newblock {\em ACM Computing Surveys}, 56(2):1--37, 2023.

\bibitem[\protect\citeauthoryear{Chen \bgroup \em et al.\egroup }{2024}]{chen2024fairCvpr}
Yuhang Chen, Wenke Huang, and Mang Ye.
\newblock {Fair Federated Learning under Domain Skew with Local Consistency and Domain Diversity}.
\newblock In {\em {IEEE/CVF Conference on Computer Vision and Pattern Recognition}}, pages 12077--12086, 2024.

\bibitem[\protect\citeauthoryear{Cui \bgroup \em et al.\egroup }{2022}]{Collab_KDD_22}
Sen Cui, Jian Liang, Weishen Pan, Kun Chen, Changshui Zhang, and Fei Wang.
\newblock {Collaboration Equilibrium in Federated Learning}.
\newblock In {\em {ACM SIGKDD Conference on Knowledge Discovery and Data Mining}}, page 241–251, 2022.

\bibitem[\protect\citeauthoryear{Dehdashtian \bgroup \em et al.\egroup }{2024}]{Dehdashtian_2024_CVPR}
Sepehr Dehdashtian, Bashir Sadeghi, and Vishnu~Naresh Boddeti.
\newblock {Utility-Fairness Trade-Offs and How to Find Them}.
\newblock In {\em {IEEE/CVF Conference on Computer Vision and Pattern Recognition}}, pages 12037--12046, 2024.

\bibitem[\protect\citeauthoryear{Du \bgroup \em et al.\egroup }{2020}]{Wei}
Wei Du, Depeng Xu, Xintao Wu, and Hanghang Tong.
\newblock {Fairness-aware Agnostic Federated Learning}.
\newblock arXiv (Preprint) arXiv:2010.05057, 2020.

\bibitem[\protect\citeauthoryear{Esipova \bgroup \em et al.\egroup }{2023}]{esipova2023disparate}
Maria~S. Esipova, Atiyeh~Ashari Ghomi, Yaqiao Luo, and Jesse~C Cresswell.
\newblock {Disparate Impact in Differential Privacy from Gradient Misalignment}.
\newblock In {\em {International Conference on Learning Representations}}, page 1–23, 2023.

\bibitem[\protect\citeauthoryear{Ezzeldin \bgroup \em et al.\egroup }{2023}]{ezzeldin}
Yahya~H Ezzeldin, Shen Yan, Chaoyang He, Emilio Ferrara, and A~Salman Avestimehr.
\newblock {Fairfed: Enabling Group Fairness in Federated Learning}.
\newblock In {\em {AAAI Conference on Artificial Intelligence}}, pages 7494--7502, 2023.

\bibitem[\protect\citeauthoryear{Hamman and Dutta}{2024}]{hamman2024demystifying_ICLR24}
Faisal Hamman and Sanghamitra Dutta.
\newblock {Demystifying Local \& Global Fairness Trade-offs in Federated Learning Using Partial Information Decomposition}.
\newblock In {\em {International Conference on Learning Representations}}, pages 1--25, 2024.

\bibitem[\protect\citeauthoryear{Huang \bgroup \em et al.\egroup }{2020}]{huang2020efficiency}
Tiansheng Huang, Weiwei Lin, Wentai Wu, Ligang He, Keqin Li, and Albert~Y Zomaya.
\newblock {An efficiency-boosting Client Selection Scheme for Federated Learning with Fairness Guarantee}.
\newblock {\em {IEEE Transactions on Parallel and Distributed Systems}}, 32(7):1552--1564, 2020.

\bibitem[\protect\citeauthoryear{Kearns \bgroup \em et al.\egroup }{2019}]{NEURIPS2019}
Michael Kearns, Aaron Roth, and Saeed Sharifi-Malvajerdi.
\newblock {Average Individual Fairness: Algorithms, Generalization and Experiments}.
\newblock In {\em {International Conference on Neural Information Processing Systems}}, pages 8242 -- 8251, 2019.

\bibitem[\protect\citeauthoryear{Lai \bgroup \em et al.\egroup }{2021}]{OORT}
Fan Lai, Xiangfeng Zhu, Harsha~V. Madhyastha, and Mosharaf Chowdhury.
\newblock {Oort: Efficient Federated Learning via Guided Participant Selection}.
\newblock In {\em {USENIX} Symposium on Operating Systems Design and Implementation}, pages 19--35, 2021.

\bibitem[\protect\citeauthoryear{Lewis \bgroup \em et al.\egroup }{2024}]{lewis2024ensuring}
Cody Lewis, Vijay Varadharajan, Nasimul Noman, and Uday Tupakula.
\newblock {Ensuring Fairness and Gradient Privacy in Personalized Heterogeneous Federated Learning}.
\newblock {\em {ACM Transactions on Intelligent Systems and Technology}}, 15(3):1--30, 2024.

\bibitem[\protect\citeauthoryear{Li \bgroup \em et al.\egroup }{2020a}]{li2020fair_ICLR}
Tian Li, Maziar Sanjabi, Ahmad Beirami, and Virginia Smith.
\newblock {Fair Resource Allocation in Federated Learning}.
\newblock In {\em {International Conference on Learning Representations}}, pages 1--27, 2020.

\bibitem[\protect\citeauthoryear{Li \bgroup \em et al.\egroup }{2020b}]{XiangLi}
Xiang Li, Kaixuan Huang, Wenhao Yang, Shusen Wang, and Zhihua Zhang.
\newblock {On the Convergence of FedAvg on Non-IID Data}.
\newblock In {\em {International Conference on Learning Representations}}, pages 1--26, 2020.

\bibitem[\protect\citeauthoryear{Li \bgroup \em et al.\egroup }{2021}]{li2021dittofairrobustfederated}
Tian Li, Shengyuan Hu, Ahmad Beirami, and Virginia Smith.
\newblock {Ditto: Fair and Robust Federated Learning Through Personalization}.
\newblock In {\em {38th International Conference on Machine Learning}}, pages 6357--6368, 2021.

\bibitem[\protect\citeauthoryear{Liu \bgroup \em et al.\egroup }{2024}]{LiuZ0HGS24}
Jiahao Liu, Yipeng Zhou, Di~Wu, Miao Hu, Mohsen Guizani, and Quan~Z. Sheng.
\newblock {FedLMT: Tackling System Heterogeneity of Federated Learning via Low-Rank Model Training with Theoretical Guarantees}.
\newblock In {\em {International Conference on Machine Learning}}, 2024.

\bibitem[\protect\citeauthoryear{Lyu \bgroup \em et al.\egroup }{2020}]{Lingjuan}
Lingjuan Lyu, Xinyi Xu, Qian Wang, and Han Yu.
\newblock Collaborative fairness in federated learning.
\newblock In {\em Federated Learning: Privacy and Incentive}, pages 189--204. Springer, 2020.

\bibitem[\protect\citeauthoryear{McMahan \bgroup \em et al.\egroup }{2017}]{mcmahan2017communication}
Brendan McMahan, Eider Moore, Daniel Ramage, Seth Hampson, and Blaise~Aguera y~Arcas.
\newblock {Communication-efficient Learning of Deep Networks from Decentralized Data}.
\newblock In {\em {Artificial Intelligence and Statistics}}, pages 1273--1282, 2017.

\bibitem[\protect\citeauthoryear{Meerza and Liu}{2024}]{ijcai2024_EABFL}
Syed Irfan~Ali Meerza and Jian Liu.
\newblock {EAB-FL: Exacerbating Algorithmic Bias through Model Poisoning Attacks in Federated Learning}.
\newblock In {\em {International Joint Conference on Artificial Intelligence}}, pages 458--466, 2024.

\bibitem[\protect\citeauthoryear{Mohri \bgroup \em et al.\egroup }{2019}]{mohri2019agnostic}
Mehryar Mohri, Gary Sivek, and Ananda~Theertha Suresh.
\newblock {Agnostic Federated Learning}.
\newblock In {\em {International Conference on Machine Learning}}, pages 4615--4625, 2019.

\bibitem[\protect\citeauthoryear{Pan \bgroup \em et al.\egroup }{2024}]{FedLF}
Zibin Pan, Chi Li, Fangchen Yu, Shuyi Wang, Haijin Wang, Xiaoying Tang, and Junhua Zhao.
\newblock {FedLF: Layer-Wise Fair Federated Learning}.
\newblock In {\em AAAI Conference on Artificial Intelligence}, pages 14527--14535, 2024.

\bibitem[\protect\citeauthoryear{Rafi \bgroup \em et al.\egroup }{2024}]{rafi2024fairness}
Taki~Hasan Rafi, Faiza~Anan Noor, Tahmid Hussain, and Dong-Kyu Chae.
\newblock {Fairness and Privacy Preserving in Federated Learning: A Survey}.
\newblock {\em {Information Fusion}}, 105(C):102198, 2024.

\bibitem[\protect\citeauthoryear{Shi \bgroup \em et al.\egroup }{2023}]{shi2023fairness}
Yuxin Shi, Zelei Liu, Zhuan Shi, and Han Yu.
\newblock {Fairness-aware Client Selection for Federated Learning}.
\newblock In {\em {IEEE International Conference on Multimedia and Expo}}, pages 324--329, 2023.

\bibitem[\protect\citeauthoryear{Shi \bgroup \em et al.\egroup }{2024a}]{Shi}
Yuxin Shi, Han Yu, and Cyril Leung.
\newblock {Towards Fairness-Aware Federated Learning}.
\newblock {\em IEEE Transactions on Neural Networks and Learning Systems}, 35(9):11922--11938, 2024.

\bibitem[\protect\citeauthoryear{Shi \bgroup \em et al.\egroup }{2024b}]{shi2022fedfaim}
Zhuan Shi, Lan Zhang, Zhenyu Yao, Lingjuan Lyu, Cen Chen, Li~Wang, Junhao Wang, and Xiang-Yang Li.
\newblock {Fedfaim: A Model Performance-based Fair Incentive Mechanism for Federated Learning}.
\newblock {\em {IEEE Transactions on Big Data}}, 10(6):1038--1050, 2024.

\bibitem[\protect\citeauthoryear{Shui \bgroup \em et al.\egroup }{2022}]{Changjian}
Changjian Shui, Gezheng Xu, Qi~CHEN, Jiaqi Li, Charles~X. Ling, Tal Arbel, Boyu Wang, and Christian Gagn\'{e}.
\newblock {On Learning Fairness and Accuracy on Multiple Subgroups}.
\newblock In {\em {International Conference on Neural Information Processing Systems}}, pages 34121--34135, 2022.

\bibitem[\protect\citeauthoryear{Soltani \bgroup \em et al.\egroup }{2023}]{ijcai2023_survey}
Behnaz Soltani, Yipeng Zhou, Venus Haghighi, and John C.~S. Lui.
\newblock {A Survey of Federated Evaluation in Federated Learning}.
\newblock In {\em { International Joint Conference on Artificial Intelligence}}, pages 6769--6777, 2023.

\bibitem[\protect\citeauthoryear{Tastan \bgroup \em et al.\egroup }{2024}]{IJCAI_Shapley}
Nurbek Tastan, Samar Fares, Toluwani Aremu, Samuel Horváth, and Karthik Nandakumar.
\newblock {Redefining Contributions: Shapley-Driven Federated Learning}.
\newblock In {\em {International Joint Conference on Artificial Intelligence}}, pages 5009--5017, 2024.

\bibitem[\protect\citeauthoryear{Wang \bgroup \em et al.\egroup }{2021}]{wang2021FedFV}
Zheng Wang, Xiaoliang Fan, Jianzhong Qi, Chenglu Wen, Cheng Wang, and Rongshan Yu.
\newblock {Federated Learning with Fair Averaging}.
\newblock In {\em {International Joint Conference on Artificial Intelligence}}, pages 1615--1623, 2021.

\bibitem[\protect\citeauthoryear{Woisetschläger \bgroup \em et al.\egroup }{2024}]{ijcai2024p919}
Herbert Woisetschläger, Alexander Erben, Shiqiang Wang, Ruben Mayer, and Hans-Arno Jacobsen.
\newblock {A Survey on Efficient Federated Learning Methods for Foundation Model Training}.
\newblock In {\em {International Joint Conference on Artificial Intelligence}}, pages 8317--8325, 2024.

\bibitem[\protect\citeauthoryear{Wu \bgroup \em et al.\egroup }{2024}]{IJCAI_24_Fairness}
Nannan Wu, Zhuo Kuang, Zengqiang Yan, and Li~Yu.
\newblock {From Optimization to Generalization: Fair Federated Learning Against Quality Shift via Inter-client Sharpness Matching}.
\newblock In {\em International Joint Conference on Artificial Intelligence}, pages 5199 -- 5207, 2024.

\bibitem[\protect\citeauthoryear{Xu and Wang}{2021}]{Xu_bandwidth}
Jie Xu and Heqiang Wang.
\newblock Client selection and bandwidth allocation in wireless federated learning networks: A long-term perspective.
\newblock {\em IEEE Transactions on Wireless Communications}, 20(2):1188--1200, 2021.

\bibitem[\protect\citeauthoryear{Yang \bgroup \em et al.\egroup }{2023}]{YangZHWS23}
Yunchao Yang, Yipeng Zhou, Miao Hu, Di~Wu, and Quan~Z. Sheng.
\newblock {BARA: Efficient Incentive Mechanism with Online Reward Budget Allocation in Cross-Silo Federated Learning}.
\newblock In {\em {International Joint Conference on Artificial Intelligence}}, 2023.

\bibitem[\protect\citeauthoryear{Yu \bgroup \em et al.\egroup }{2020}]{yu}
Han Yu, Zelei Liu, Yang Liu, Tianjian Chen, Mingshu Cong, Xi~Weng, Dusit Niyato, and Qiang Yang.
\newblock {A Fairness-aware Incentive Scheme for Federated Learning}.
\newblock In {\em AAAI/ACM Conference on AI, Ethics, and Society}, page 393–399, 2020.

\bibitem[\protect\citeauthoryear{Zhang \bgroup \em et al.\egroup }{2021}]{Incentive_WWW21}
Jingwen Zhang, Yuezhou Wu, and Rong Pan.
\newblock {Incentive Mechanism for Horizontal Federated Learning Based on Reputation and Reverse Auction}.
\newblock In {\em {The ACM Web Conference 2021}}, page 947–956, 2021.

\bibitem[\protect\citeauthoryear{Zhang \bgroup \em et al.\egroup }{2022}]{AAAImodelperformance}
Sai~Qian Zhang, Jieyu Lin, and Qi~Zhang.
\newblock {A Multi-agent Reinforcement Learning Approach for Efficient Client Selection in Federated Learning}.
\newblock In {\em {AAAI Conference on Artificial Intelligence}}, pages 9091--9099, 2022.

\bibitem[\protect\citeauthoryear{Zhang \bgroup \em et al.\egroup }{2023}]{NEURIPS2023_2e0d3c6a}
Jianqing Zhang, Yang Hua, Jian Cao, Hao Wang, Tao Song, Zhengui XUE, Ruhui Ma, and Haibing Guan.
\newblock {Eliminating Domain Bias for Federated Learning in Representation Space}.
\newblock In {\em {International Conference on Neural Information Processing Systems}}, pages 14204--14227, 2023.

\bibitem[\protect\citeauthoryear{Zhang \bgroup \em et al.\egroup }{2024}]{ACM2024privacy}
Yanjun Zhang, Ruoxi Sun, Liyue Shen, Guangdong Bai, Minhui Xue, Mark~Huasong Meng, Xue Li, Ryan Ko, and Surya Nepal.
\newblock {Privacy-preserving and Fairness-aware Federated Learning for Critical Infrastructure Protection and Resilience}.
\newblock In {\em {The ACM Web Conference 2024}}, pages 2986--2997, 2024.

\bibitem[\protect\citeauthoryear{Zhao \bgroup \em et al.\egroup }{2022}]{zhao2022participant}
Jianxin Zhao, Xinyu Chang, Yanhao Feng, Chi~Harold Liu, and Ningbo Liu.
\newblock {Participant Selection for Federated Learning with Heterogeneous Data in Intelligent Transport System}.
\newblock {\em {IEEE Transactions on Intelligent Transportation Systems}}, 24(1):1106--1115, 2022.

\bibitem[\protect\citeauthoryear{Zhu \bgroup \em et al.\egroup }{2023}]{CVPR_Personlization}
Junyi Zhu, Xingchen Ma, and Matthew~B. Blaschko.
\newblock {Confidence-Aware Personalized Federated Learning via Variational Expectation Maximization}.
\newblock In {\em {IEEE/CVF Conference on Computer Vision and Pattern Recognition)}}, pages 24542--24551, 2023.

\end{thebibliography}
\end{document}